\documentclass[runningheads]{llncs}

\usepackage{mathtools}
\usepackage{amssymb}
\usepackage{amsmath}

\usepackage{pdfpages}
\usepackage{booktabs}
\usepackage{graphicx}
\usepackage{url}
\usepackage{tikz}
\usepackage{enumitem}
\usepackage{dirtytalk}
\usepackage{xcolor}
\usepackage{comment}

\newcommand{\Fig}[1]{Fig.~#1}
\newcommand{\Tab}[1]{Tab.~#1}
\newcommand{\Sec}[1]{Sec.~#1}

\newcommand{\Eq}[1]{Eq.~#1}

\newcommand{\myDescription}[1]{
    \vspace{.5em}
    \begin{itemize}[leftmargin=0mm,label={}]
        \setlength\itemsep{.5em}
        #1
    \end{itemize}
}
\newcommand{\myItem}[2]{
    \item \textbf{#1}\: #2
}

\def \gridworld {8-GridWorld}
\def \cheb {Chebyshev}
\def \bbastar {Black-Box A*}

\def \nwastar {Neural Weighted A*}

\def \rawbba {BBA*}
\def \rawna {NA*}
\def \rawadmna {ADM\_NA*}
\def \rawnsna {NS\_NA*}
\def \rawnwa {NWA*}
\def \bba {\emph{\rawbba}}
\def \na {\emph{\rawna}}
\def \admna {\emph{\rawadmna}}
\def \nsna {\emph{\rawnsna}}
\def \nwa {\emph{\rawnwa}}

\newcommand{\meanstd}[2]{$#1$}
\newcommand{\meanstdbf}[2]{$\mathbf{#1}$}
\newcommand{\metrics}{CR & Gen. CR & EN & Gen. EN}

\title{\nwastar: Learning Graph Costs and Heuristics with Differentiable Anytime A*}
\titlerunning{\nwastar}

\author{Alberto Archetti\orcidID{0000-0003-3826-4645} \and Marco Cannici\orcidID{0000-0002-9217-3552} \and Matteo Matteucci\orcidID{0000-0002-8306-6739}}

\institute{Politecnico di Milano, Milano 20133, Italy\\
\email{alberto1.archetti@mail.polimi.it}\\
\email{\{marco.cannici,matteo.matteucci\}@polimi.it}}

\authorrunning{A. Archetti, M. Cannici, and M. Matteucci}

\def \dataseturl{\url{https://github.com/archettialberto/tilebased_navigation_datasets}}

\def \sourceurl{\url{https://github.com/archettialberto/neural_weighted_a_star}}

\begin{document}

\maketitle

\begin{abstract}
Recently, the trend of incorporating differentiable algorithms into deep learning architectures arose in machine learning research, as the fusion of neural layers and algorithmic layers has been beneficial for handling combinatorial data, such as shortest paths on graphs. Recent works related to data-driven planning aim at learning either cost functions or heuristic functions, but not both. We propose \nwastar, a differentiable anytime planner able to produce improved representations of planar maps as graph costs and heuristics. Training occurs end-to-end on raw images with direct supervision on planning examples, thanks to a differentiable A* solver integrated into the architecture. More importantly, the user can trade off planning accuracy for efficiency at run-time, using a single, real-valued parameter. The solution suboptimality is constrained within a linear bound equal to the optimal path cost multiplied by the tradeoff parameter. We experimentally show the validity of our claims by testing \nwastar\ against several baselines, introducing a novel, tile-based navigation dataset. We outperform similar architectures in planning accuracy and efficiency.
\end{abstract}

\keywords{Weighted A*\and Differentiable algorithms \and Data-based planning.}

\section{Introduction}
\label{sec:intro}

A*~\cite{a_star} is the most famous heuristic-based planning algorithm, and it constitutes one of the essentials for the computer scientist's toolbox. It is widely used in robotic motion~\cite{arm_manipulation} and navigation systems~\cite{urban_navigation}, but its range extends to all the fields that benefit from shortest path search on graphs~\cite{survey_a_star}. Differently from other shortest path algorithms, such as Dijkstra~\cite{dijkstra} or Greedy Best First~\cite{russelnorvig}, A* is known to be optimally efficient~\cite{russelnorvig}. This means that, besides returning the optimal solution, there is no other algorithm that can be more efficient, in general, provided the same admissible heuristic. % function. 
Even though optimality seems a desirable property for A*, it is often more of a burden than a virtue in practical applications. This is because, in the worst case, A* takes exponential time to converge to the optimal solution, and this is not affordable in large search spaces. 

%Another compelling issue of A* planning is that hand-crafting non-trivial heuristic functions is costly and reliant on domain
Another compelling issue of A* is that hand-crafting non-trivial heuristics is costly and reliant on domain knowledge. Despite the prolific research in deep-learning-based graph labeling~\cite{dl_on_graphs}, neural networks often struggle with data exhibiting combinatorial complexity, such as shortest paths~\cite{black_box}. For this reason, many researchers started including differentiable algorithmic layers directly into deep learning pipelines. These layers implement algorithms with combinatorial operations in the forward pass, while providing a smooth, approximated derivative in the backward pass. This approach helps the neural components to converge faster with fewer data samples, promoting the birth of hybrid architectures, trainable end-to-end, that extend the reach of deep learning to complex combinatorial problems. Many backpropagation-ready algorithmic layers have been developed, such as~\cite{optnet,amos2019differentiable,berthet2020learning,black_box,wang2019satnet,neural_a_star}. Among these, some~\cite{berthet2020learning,black_box,neural_a_star} propose differentiable shortest path solvers able to learn graph costs from planning examples on raw image inputs. However, none of the previous works tackles heuristic design, which is the essential aspect that makes A* scale to complex scenarios.

\begin{figure}[t]
    \centering
    \includegraphics[width=0.85\textwidth]{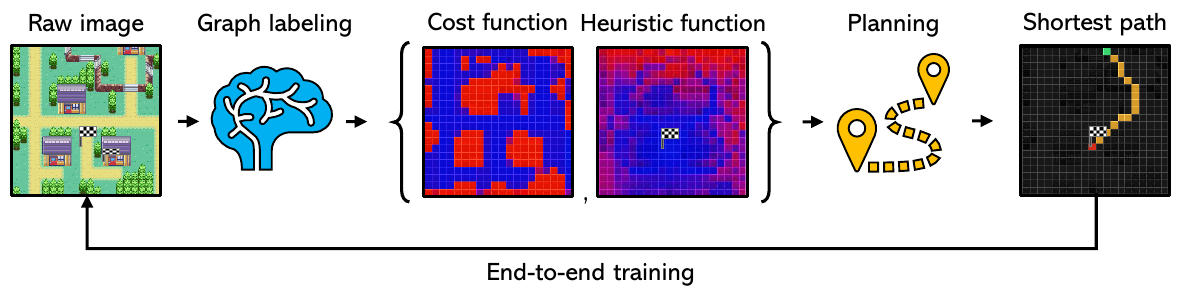}
    \caption{Can we learn to navigate a terrain effectively by just looking at its map? Neural Weighted A* accurately predicts from the raw image of the navigation area the costs of traversing local regions and a global heuristic for reaching the destination.}
    \label{fig:wh_example}
\end{figure}

With Neural Weighted A* (NWA*), we propose the first deep-learning-based differentiable planner able to predict graph costs and heuristic functions from unlabeled images of navigation areas (\Fig{\ref{fig:wh_example}}). Training occurs end-to-end on shortest-path examples, exploiting a fusion of differentiable planners from~\cite{black_box,neural_a_star}. Also, NWA* is the first architecture that enables the user to trade off planning accuracy for convergence speed with a single, real-valued parameter, even at runtime. Balancing search accuracy and efficiency from unlabeled images is crucial in many navigation problems. Among the most notable examples, we find hierarchical planning for robotic navigation, where accuracy and efficiency assume a different priority depending on the spacial granularity at which planning is executed, and real-world pedestrian modeling, where graph labeling from raw images is not feasible by hand~\cite{neural_a_star}. As a final remark, since our method arises from the Weighted A* algorithm~\cite{anytime_heuristic_search}, the solution cost never exceeds the optimal one by a factor proportional to the tradeoff parameter. 

We extensively test Neural Weighted A* against the baselines from~\cite{black_box,neural_a_star}, and conduct experiments on two tile-based datasets. The first is adapted from~\cite{black_box}, while the second dataset is novel, and its goal is to provide a scenario more complex than the first. Both datasets are publicly available (\Sec{\ref{sec:datasets}}). In summary, with Neural Weighted A*, we make the following contributions:
\begin{itemize}
    \item We develop the first deep-learning system able to generate both cost functions and heuristic functions in a principled way from raw map images.
    \item We propose the first method to smoothly trade off planning accuracy and efficiency at runtime, compliant with the Weighted A* bound on solution suboptimality.
    \item We augment an existing dataset and propose a new one for planning benchmarks on planar navigation problems.
\end{itemize}

\section{Related Work}
\label{sec:rel_works}

Connections between deep learning and differentiable algorithms arise from different domains. The first examples lie within the 3D rendering literature~\cite{kato2020differentiable}, composing neural-network-based encoders with differentiable renderer-like decoders to learn the constructing parameters of the input scene. Differentiable decoders spread to physics simulations~\cite{NIPS2018_7948,seo2019differentiable}, logical reasoning~\cite{wang2019satnet,Yang2017DifferentiableLO}, and control~\cite{amos2019differentiable,east2020infinitehorizon,karkus2019differentiable}. Combinatorial optimization is also a topic of interest, from differentiable problem-specific solvers~\cite{bello2017neural,nazari2018reinforcement,wang2019satnet} to general-purpose ones~\cite{optnet,berthet2020learning,black_box}. Indeed, many combinatorial algorithms and their differentiable implementation have already been studied, such as Traveling Salesman~\cite{bello2017neural,Deudon2018LearningHF}, (Conditional) Markov Random Fields~\cite{chen2015learning,liu2015semantic,paschalidou2019raynet,Zheng_2015}, and Shortest Path~\cite{neural_a_star}. Each of these works shows how structured differentiable components enable deep learning architectures to learn combinatorial patterns easily from data.

A handful of works started experimenting with convexity, one of the most important properties of combinatorial optimization. The first is the neural layer by Amos et al.~\cite{amos2017input}, which is constrained to learn convex functions only. Following this work, Pitis et al.~\cite{pitis2020inductive} exploit the convex neural layer to design a trainable graph-embedding metric that respects triangle inequality. This is beneficial for learning graph costs that encode mathematically sound distances, even though the method is limited to train on a single graph.

Among search-based planning research, some studies focus on a data-driven approach where planning cues are inferred from raw image inputs~\cite{berthet2020learning,black_box,neural_a_star}. In~\cite{black_box}, Vlastelica et al. develop a technique to differentiate solvers for integer linear optimization problems, treating them as black-boxes. As the shortest path belongs to this set of problems~\cite{ilpformulation}, the authors are able to map images of navigation areas to extremely accurate graph costs, such that the paths evaluated on the cost predictions closely resembles the ground-truth ones. At its core, the technique from~\cite{black_box} consists of a smooth interpolation of the piecewise constant function defined by the black-box solver. This technique is well suited for learning accurate costs, but, due to its black-box nature, cannot address heuristic design, the aspect that makes the search efficient, and which we study in this work.

On the other side of the spectrum, Yonetani et al.~\cite{neural_a_star}, reformulate the canonical A* algorithm as a set of differentiable tensor operations. Their goal is to develop a deep-learning-based architecture able to learn improved cost functions such that the planning search avoids non-convenient regions to traverse. In order to train the architecture, the authors define a loss function that minimizes the difference between the nodes expanded by A* and the ground truth paths. Therefore, the neural network is forced to learn shortcuts and bypasses that severely accelerate the search, but may result in inaccurate path predictions.

Lastly, Berthet et al.~\cite{berthet2020learning} propose a general-purpose differentiable combinatorial optimizer based on stochastic perturbations with a strong theoretical insight. Despite this work being more recent and general than~\cite{black_box}, it was outperformed by~\cite{black_box} in our experimental settings. Therefore, we choose to focus on~\cite{black_box} for the rest of the paper. In fact, our work builds on~\cite{black_box,neural_a_star} to develop the first learning architecture that is not forced to choose to plan either accurately or efficiently, but is able to smoothly tradeoff between these opposing aspects of planning. 

%%%%%%%%%%%%%%%%%%%%%%%%%%%%%%%%%%%%%%%%%%%%%%%%%%%%%%%%%%%%%%%%%%%%%%%%%%%%%%%%
\section{Preliminaries}
% \subsection{Graphs and GridWorld}

Let $\mathcal{G} = (\mathcal{N}, \mathcal{E})$ be a graph where $\mathcal{N}$ is a finite set of nodes and $\mathcal{E}$ is a finite set of edges connecting the nodes. Let $s$ and $t$ be two distinct nodes from $\mathcal{N}$, called source and target. We define a path $y$ on $\mathcal{G}$ connecting $s$ to $t$ as a sequence of adjacent nodes $(n_0, n_1, \dots, n_k)$ such that $n_0 = s$, $n_k = t$, and each node is traversed at most once. %We call $\mathcal{Y}_{st}$ the set of all paths connecting $s$ and $t$ on $\mathcal{G}$.

In this work, we always refer to the \gridworld\ setting~\cite{berthet2020learning,data_driven_planning,black_box,neural_a_star} but the techniques we describe can be easily applied to general graph settings, nevertheless. In \gridworld, nodes are disposed in a grid-like pattern, and edges connect only the nodes belonging to neighboring cells, including the diagonal ones. Each node is paired with a non-negative, real-valued cost belonging to a cost function $\bar{W} \in \mathbb{R}_+^\mathcal{N}$. Paths are represented in binary form as $Y \in \{0, 1\}^\mathcal{N}$ with ones corresponding to the traversed nodes. The total cost of a path $Y$, denoted as $\langle \bar{W}, Y \rangle$, is the sum of its nodes' costs. Given a graph $\mathcal{G}$ with costs $\bar{W}$, a source node $s$, and a target node $t$, the shortest path problem consists of finding the path $\bar{Y}$ having the minimal total cost among all the paths connecting $s$ and $t$.

%%%%%%%%%%%%%%%%%%%%%%%%%%%%%%%%%%%%%%%%%%%%%%%%%%%%%%%%%%%%%%%%%%%%%%%%%%%%%%%%
\subsection{A* and Weighted A*}
\label{sec:weighted_astar}

We focus on A*~\cite{a_star}, a heuristic-based shortest path algorithm for graphs. A* searches for a minimum-cost path from $s$ to $t$ iteratively expanding nodes according to the priority measure
\begin{equation}
    F(n) = G(n) + H(n)\text{.}
    \label{eq:a_star}
\end{equation}
$G(n)$ is the exact cumulated cost from $s$ to $n$, and $H(n)$ is a heuristic function estimating the cost between $n$ and $t$. A* is known to be optimally efficient when $H(n)$ is admissible~\cite{russelnorvig}, i.e., it never overestimates the optimal cost between $n$ and $t$. An example of admissible heuristic on \gridworld\ is
\begin{equation}
    H_C(n) = w_{\text{min}} \cdot D_C(n, t)
    \label{eq:cheb_h}
\end{equation}
where $w_{\text{min}} = \min_{n \in \mathcal{N}} \bar{W}(n)$ and $D_C(n, t)$ is the \cheb\ distance between $n$ and $t$ in the grid, i.e., $D_C(n, t) = \max\{|n_x - t_x|, |n_y - t_y|\}$.

For large graphs, A* may take exponential time to find the optimal solution~\cite{anytime_heuristic_search}. Hence, in practical applications, it is preferable to find an approximate solution quickly, sacrificing the optimality constraint. This idea is explored by one of A*'s extensions, called Weighted A* (WA*)~\cite{semiadmh}. This algorithm is equivalent to a standard A* search, but the heuristic $H(n)$ in \Eq{\ref{eq:a_star}} is scaled up by a factor of $1 + \epsilon$, where $\epsilon \geq 0$. Assuming $H(n)$ to be admissible, WA* returns the optimal path for $\epsilon = 0$. Conversely, for $\epsilon > 0$, the heuristic function drives the search, leading to fewer node expansions, but influencing the path trajectory. The cost difference between the WA* solution $Y$ and the optimal path $\bar{Y}$ is linearly bounded~\cite{anytime_heuristic_search}:
\begin{equation}
    \langle \bar{W}, Y \rangle \leq (1 + \epsilon) \cdot \langle \bar{W}, \bar{Y} \rangle\text{.}
    \label{eq:w_a_star_bound}
\end{equation}

\section{Neural Weighted A*}

\begin{figure}[t]
    \centering
    \includegraphics[width=.9\textwidth]{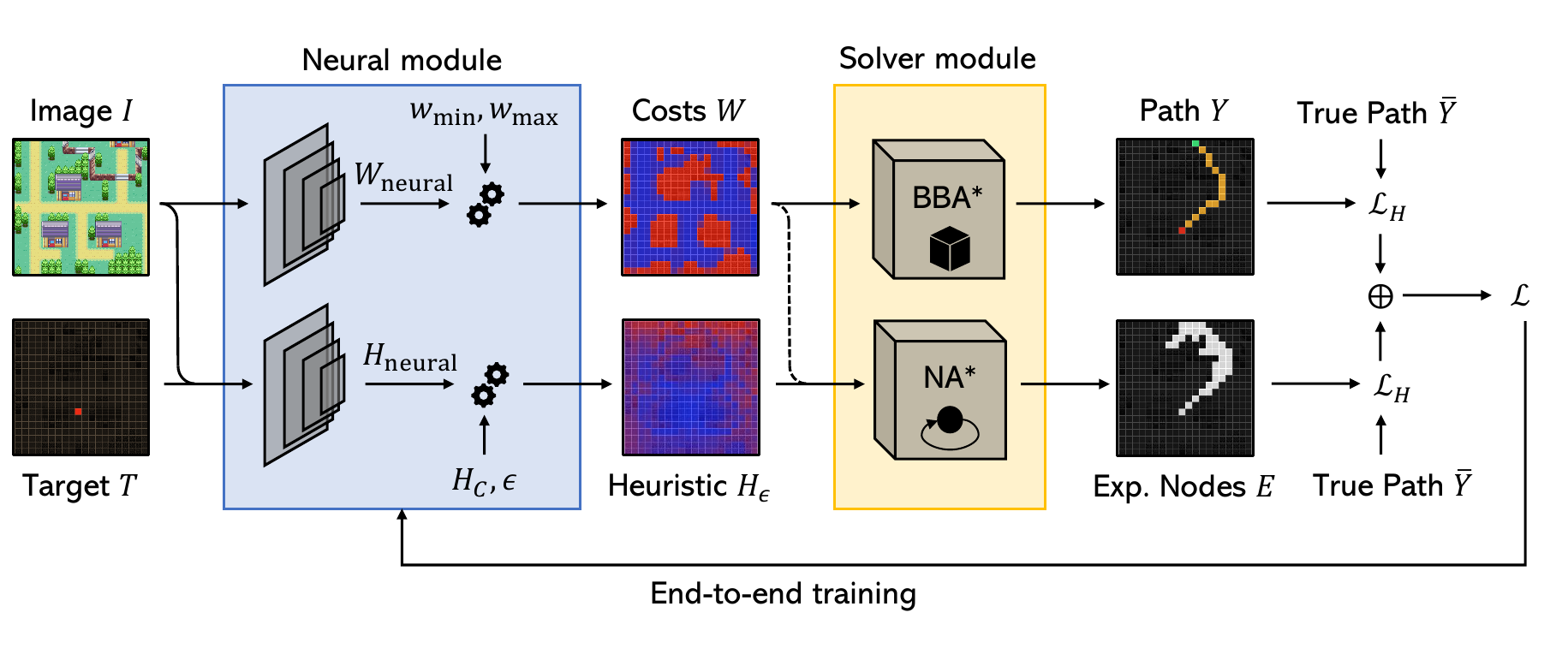}
    \caption{Schematics of Neural Weighted A*. The neural module (blue) predicts the costs $W$ and the heuristic function $H_\epsilon$. The solver module (yellow) runs two A* solvers, differentiable according to the techniques described in~\cite{black_box,neural_a_star}. The first solver computes the shortest path $Y$, while the second computes the nodes expanded by A*, $E$. 
    %The system trains end-to-end using the sum of two Hamming loss functions $\mathcal{L}_H$ evaluated with respect to the ground-truth path $\bar{Y}$.
    }
    \label{fig:nwastar}
\end{figure}

In the A* algorithm, the nodes are expanded according to the balance between the cumulated costs $G \in \mathbb{R}_+^\mathcal{N}$ and the heuristic function $H$ (\Eq{\ref{eq:a_star}}). If $G \gg H$, A* expands nodes mostly according to $G$, behaving similarly to Dijkstra's algorithm. If, on the other hand, $H \gg G$, then the A* behavior is closer to a Greedy Best First search. Therefore, tuning the scale of the final cost and heuristic functions is the key to control the tradeoff between planning accuracy and efficiency. We propose to accomplish this goal with a novel, deep-learning-based architecture for graph labeling from planning examples, called Neural Weighted A* (\Fig{\ref{fig:nwastar}}). It is composed of two modules: the \emph{neural module} (\Sec{\ref{sec:neural_module}}) and the \emph{solver module} (\Sec{\ref{sec:solver_module}}). The neural module generates planning-ready graph costs and a heuristic function from the top view of the navigation area. The solver module executes the planning procedure in the forward pass, while providing a smooth derivative in the backward pass to enable end-to-end training.

In the following, we indicate with the \say{neural} subscript values directly coming out of neural networks, such as $W_\text{neural}$ and $H_\text{neural}$, while we use the bar superscript, as in $\bar{W}$ and $\bar{Y}$, to indicate ground-truth values. 

%We establish the following goals for the Neural Weighted A* architecture:
%\begin{itemize}
%    \item Predict a cost function $W$ and a source-agnostic heuristic function $H_\epsilon$ from a raw image input such that planning on these functions results in a path prediction $Y$ as close as possible to the ground-truth $\bar{Y}$.
%    \item Train the system end-to-end with supervision on shortest path examples, such that we can infer planning information from raw map images.
%    \item Design the heuristic function $H_\epsilon$ such that it can gradually trade off planning accuracy for faster convergence speed. Tuning occurs using a real-valued parameter, $\epsilon \geq 0$, even at runtime.
%    \item When lowering accuracy for faster planning, the returned path cost never exceeds the optimal one by a factor proportional to $\epsilon$.
%\end{itemize}

%%%%%%%%%%%%%%%%%%%%%%%%%%%%%%%%%%%%%%%%%%%%%%%%%%%%%%%%%%%%%%%%%%%%%%%%%%%%%%%%
\subsection{The neural module}
\label{sec:neural_module}

The neural module is composed of two fully-convolutional neural networks. The first one (upper network in \Fig{\ref{fig:nwastar}}) processes a color image $I \in [0, 1]^{\Gamma \times 3}$ of resolution $\Gamma$ returning a cost prediction $W_\text{neural} \in [0, 1]^\mathcal{N}$. The second neural network (lower network in \Fig{\ref{fig:nwastar}}) takes as input the concatenation of $I$ and the target $T$, i.e., a matrix with a one corresponding to the target position scaled up to the image resolution $\Gamma$, and returns a heuristic prediction $H_\text{neural} \in [0, 1]^\mathcal{N}$. This separation enforces the system to learn costs that are target-agnostic, since $T$ is not included in the input of the first neural network. 

%In the A* algorithm, the nodes are expanded according to the balance between the cumulated costs $G \in \mathbb{R}_+^\mathcal{N}$ and the heuristic function $H$ (\Eq{\ref{eq:a_star}}). If $G \gg H$, A* expands nodes mostly according to $G$, behaving similarly to Dijkstra's algorithm. If, on the other hand, $H \gg G$, then the A* behavior is closer to a Greedy Best First search. Therefore, tuning the scale of the final cost and heuristic functions is the key to control the tradeoff between planning accuracy and efficiency. To do so, first, 

In order to control the relative magnitude between the costs and the heuristic function, we uniformly scale the values of $W_\text{neural}$ in the interval $[w_{\text{min}}, w_{\text{max}}]$ such that $w_{\text{min}} > 0$. We call $W$ the new and final cost function. Then, we compute the final heuristic function as 
\begin{equation}
    H_\epsilon = (1 + \epsilon \cdot H_\text{neural}) \cdot H_C
    \label{eq:h_epsilon}
\end{equation}
where $H_C$ is the \cheb\ heuristic (\Eq{\ref{eq:cheb_h}}), and $\epsilon \geq 0$ is the accuracy-efficiency tradeoff parameter. 

\begin{figure}[t]
    \centering
    \includegraphics[width=.7\textwidth]{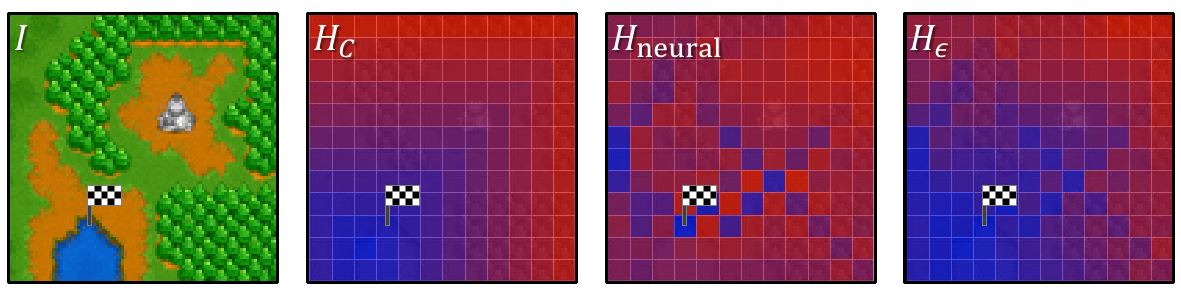}
    \caption{From left to right, first: image sample from the Warcraft dataset (\Sec{\ref{sec:datasets}}). The target node is in the bottom left region of the map. Second: \cheb\ heuristic $H_C$ (\Eq{\ref{eq:cheb_h}}). Red indicates high values; blue indicates low values. Third: neural prediction $H_\text{neural}$. Fourth: final heuristic $H_\epsilon$ (\Eq{\ref{eq:h_epsilon}}) for $\epsilon = 9$.}
    \label{fig:heuristic_comparison}
\end{figure}

For any node $n \in \mathcal{N}$, the purpose of $\epsilon$ and $H_\text{neural}(n)$ is to modulate the intensity of the final heuristic $H_\epsilon(n)$ between two values, $H_C(n)$ and $(1 + \epsilon) \cdot H_C(n)$. When $\epsilon = 0$, $H_\epsilon(n)$ is equal to the admissible \cheb\ heuristic $H_C(n)$. Therefore, the solution optimality is guaranteed. Conversely, when $\epsilon > 0$, $H_\epsilon(n)$ is not admissible, in general, anymore. However, if $n$ is a node likely to be convenient to traverse, it is mapped to a value close to $H_C(n)$, as the neural network learns to predict a value $H_\text{neural}(n) \approx 0$. If, on the other hand, $n$ seems very unlikely to be traversed, its heuristic value is scaled up by a factor of $1 + \epsilon$, as $H_\text{neural} \approx 1$. In this way, by increasing $\epsilon$, we increase the difference in heuristic values between nodes convenient and non-convenient to expand according to the neural prediction, forcing A* to prefer the nodes where $H_\text{neural} \approx 0$. \Fig{\ref{fig:heuristic_comparison}} visually illustrates the relationship between $H_C$, $H_\text{neural}$, and $H_\epsilon$.

Lastly, we observe that $H_\epsilon(n) \leq (1 + \epsilon) \cdot H_C(n)$. Since $H_C$ is an admissible heuristic function for \gridworld, we are guaranteed, by the Weighted A* bounding result (\Sec{\ref{sec:weighted_astar}}, \Eq{\ref{eq:w_a_star_bound}}), to never return a path whose cost exceeds the optimal one by a factor of $1 + \epsilon$.

%%%%%%%%%%%%%%%%%%%%%%%%%%%%%%%%%%%%%%%%%%%%%%%%%%%%%%%%%%%%%%%%%%%%%%%%%%%%%%%%
\subsection{The solver module}
\label{sec:solver_module}

The solver module is composed of two differentiable A* solvers. The first, called \bbastar, implements the A* algorithm with black-box differentiation as in~\cite{black_box}. It computes the shortest path $Y$ given the costs $W$, the admissible \cheb\ heuristic $H_C$ (\Eq{\ref{eq:cheb_h}}) and the source-target nodes. The second solver implements Neural A*, as in~\cite{neural_a_star}. It returns the nodes $E$ expanded during the A* search given $W$, $H_\epsilon$ (\Eq{\ref{eq:h_epsilon}}), and the source-target nodes. The two solvers provide two separate gradient signals. As $Y$ is computed following the black-box derivative from~\cite{black_box}, its value is differentiable only with respect to $W$. Following the Neural A* approach~\cite{neural_a_star}, instead, the matrix of expanded nodes, $E$, is differentiable only with respect to $H_\epsilon$. Within the solver module, we effectively combine the two differentiation techniques, enabling the neural module to learn both costs and heuristics with the proper gradient signal.
%$Y$ is differentiable only with respect to $W$, as $H_\epsilon$ may affect the shortest path for $\epsilon > 0$. Therefore, $H_\epsilon$ is not included in the computation of $Y$, as the differentiation technique from~\cite{black_box} requires $Y$ to be the exact result of an integer linear optimization problem with parameters $W$. The matrix of expanded nodes, $E$, is differentiable only with respect to $H_\epsilon$, instead. 
To this end, we stop propagating the gradient of $H_\epsilon$ towards $W$ in the computational graph while running the Neural A* solver (dashed arrow in \Fig{\ref{fig:nwastar}}). This is because $H_\epsilon$ is evaluated considering the target $T$, and we want to be sure that $T$ has no influence whatsoever on the target-agnostic costs $W$. 

In principle, having two separate solvers for the evaluation of $Y$ and $E$ may lead to inconsistencies, as $H_\epsilon$, for $\epsilon > 0$, affects the trajectory of the shortest path. In such a case, $Y$ may contain nodes not belonging to $E$. However, this side-effect is unavoidable during training to guarantee the correct gradient information propagation and to ensure that $W$ does not depend on $T$. These theoretical reasons are confirmed by a much lower performance during the experiments when trying to include $H_\epsilon$ as heuristic function in \bbastar. At testing time, to guarantee the output consistency, the solver module is substituted by a standard A* algorithm with $W$, $H_\epsilon$, and the source-target pair as inputs, returning $Y$ and $E$ in a single execution.

%%%%%%%%%%%%%%%%%%%%%%%%%%%%%%%%%%%%%%%%%%%%%%%%%%%%%%%%%%%%%%%%%%%%%%%%%%%%%%%%
\subsection{Loss function}

The only label required for training Neural Weighted A* is the ground-truth path $\bar{Y} \in \{0, 1\}^\mathcal{N}$. In the ideal case, both $Y$ and $E$ are equivalent to $\bar{Y}$, meaning that A* expanded only the nodes belonging to the true shortest path. 
In a more realistic case, $Y$ is close to $\bar{Y}$ following the same overall course but with minor node differences, while $E$ contains $\bar{Y}$ alongside some nodes from the surrounding area. 
Since all of these tensors contain binary values, we found the Hamming loss $\mathcal{L}_H$, as in~\cite{black_box}, to be the most effective to deal with our learning problem. The final loss is 
\begin{equation}
    \mathcal{L} = \alpha \cdot \mathcal{L}_H(\bar{Y}, Y) + \beta \cdot \mathcal{L}_H(\bar{Y}, E)
    \label{eq:loss}
\end{equation}
where $\alpha$ and $\beta$ are positive, real-valued parameters that bring the loss components to the same order of magnitude. A possible alternative to the Hamming loss is the L1 loss, as in~\cite{neural_a_star}. However, we did not find any reason to prefer it over the Hamming loss. The behavior of L1 in terms of gradient propagation is similar, but the experimental results were worse. % with respect to the Hamming loss.
\section{Data Generation}
\label{sec:datasets}

To experimentally test our claims about Neural Weighted A*, we use two tile-based datasets.\footnote{\dataseturl}
The first is a modified version of the Warcraft II dataset from~\cite{black_box} (\Sec{\ref{sec:warcraft_data}}).
%Differently from the original version, we randomly sampled the source-target nodes to make the dataset more challenging. 
The second is a novel dataset from the FireRed-LeafGreen Pokémon tileset (\Sec{\ref{sec:pkmn_data}}). In the latter, the search space is bigger, and the tileset is richer, making the setting more complex. However, Neural Weighted A* shows similar performance in both scenarios outperforming the baselines of~\cite{black_box,neural_a_star}. 
\Tab{\ref{tab:datasets_stats}} collects the summary statistics of the two datasets.

\begin{table}[t]
    \centering
    \caption{Datasets' summary statistics.}
    \begin{tabular}{@{}lll@{}}
        \toprule
        & Warcraft & Pokémon \\ 
        \midrule
        \multicolumn{1}{l}{Maps $I$ (train, validation, test)} & $10000, 1000, 1000$ & $3000, 500, 500$ \\ 
        \multicolumn{1}{l}{Map resolution $\Gamma$} & $96 \times 96$ & $320 \times 320$ \\ 
        \multicolumn{1}{l}{Tile resolution} & $8 \times 8$ & $16 \times 16$ \\ 
        \multicolumn{1}{l}{Grid shape $\mathcal{N}$} & $12 \times 12$ & $20 \times 20$ \\ 
        \multicolumn{1}{l}{Cost range} & $[0.8, 9.2]$ & $[1.0, 25.0]$ \\ 
        \multicolumn{1}{l}{Targets per image} & $2$ & $2$ \\ 
        \multicolumn{1}{l}{Sources per target} & $2$ & $2$ \\ 
        \multicolumn{1}{l}{Total number of samples} & $48000$ & $16000$ \\
        %\multicolumn{1}{l}{Path length (avg $\pm$ std)} & $11.2 \pm 2.45$ & $16.71 \pm 3.07$ \\ 
        %\multicolumn{1}{l}{Path costs (avg $\pm$ std)} & $22.66 \pm 12.83$ & $38.94 \pm 30.09$ \\ 
        \bottomrule
    \end{tabular}
    \label{tab:datasets_stats}
\end{table}

%%%%%%%%%%%%%%%%%%%%%%%%%%%%%%%%%%%%%%%%%%%%%%%%%%%%%%%%%%%%%%%%%%%%%%%%%%%%%%%%
\subsection{The Warcraft dataset}
\label{sec:warcraft_data}

\begin{figure}[t]
    \centering
    \includegraphics[width=.7\textwidth]{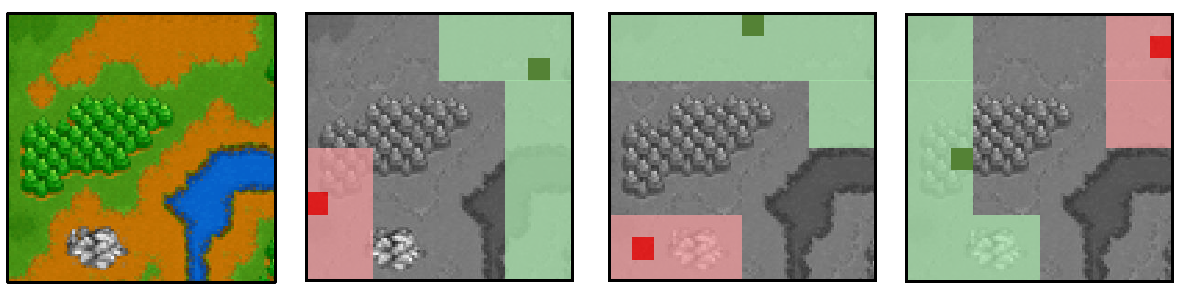}
    \caption{From left to right, first: image sample from the Warcraft dataset. Second to fourth: examples of valid source-target pairs. Red indicates the target sampling regions, while green indicates the source sampling regions.}
    \label{fig:warcraft_sampling}
\end{figure}

The original version of the Warcraft dataset~\cite{black_box} contains paths only from the top-left corner to the bottom-right corner of the image. To make the dataset more challenging, we randomly sampled the source-target pairs from $\mathcal{N}$. For each image-costs pair in the dataset, we chose two target points. The targets lie within a $3$-pixel margin from the grid edges. Then, we randomly picked two source points for each target, making four source-target pairs for each map. Each source point is sampled from the quadrant opposite to its target to ensure that each path traverses a moderate portion of the map, as shown in \Fig{\ref{fig:warcraft_sampling}}.

%%%%%%%%%%%%%%%%%%%%%%%%%%%%%%%%%%%%%%%%%%%%%%%%%%%%%%%%%%%%%%%%%%%%%%%%%%%%%%%%

\subsection{The Pokémon dataset}
\label{sec:pkmn_data}

\begin{figure}[t]
    \centering
    \includegraphics[width=.7\textwidth]{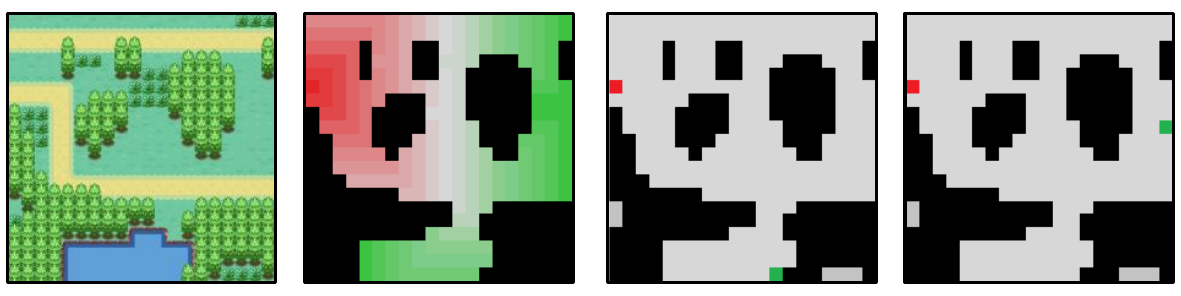}
    \caption{From left to right, first: image from the Pokémon dataset. Second: wall regions (black), number of steps from the target region (red-to-green gradient), and source-sampling region (green). Third and fourth: examples of valid source-target pairs.}
    \label{fig:pkmn_sampling}
\end{figure}

The Pokémon dataset is a novel, tile-based dataset we present in this paper. It comes with $4000$ RGB images of $320 \times 320$ pixels generated from Cartographer~\cite{cartographer_alpha}, a random Pokémon map generation tool. Each image is composed of $400$ tiles, each of $16 \times 16$ pixels, arranged in a $20 \times 20$ pattern. Each tile is linked to a real-valued cost in the interval $[1, 25]$. The training set comprises $3000$ image-costs pairs, while the test and validation sets contain $500$ pairs each. For each image-costs pair, we sampled two target points, avoiding non-traversable regions in the original Pokémon game, i.e., where $\bar{W}(n) = 25$. We refer to these regions as walls. Then, we sampled two sources for each target, such that the number of steps separating them is at least $12$ (\Fig{\ref{fig:pkmn_sampling}}). 

The Pokémon dataset provides a setting more challenging than Warcraft. First, the number of rows and columns increases from $12$ to $20$, making the search space nearly four times bigger. Also, the tileset is richer. Warcraft is limited to only five terrain types (grass, earth, forest, water, and stone), and there is a one-to-one correspondence between terrain types and cost values. These aspects make the tile-to-cost patterns very predictable for the neural component of the architectures. Pokémon, on the other hand, has double the number of individual cost values, spread between the tilesets from four different biomes: beach, forest, tundra, and desert. Also, each image sample may contain buildings. The variability in terms of visual features is richer, and similar costs may correspond to tiles exhibiting very different patterns.

\section{Experimental Validation}

In the following, we describe the experiments to test the validity of our claims. Each time we refer to results obtained with Neural Weighted A*, we note the $\epsilon$ value used for the evaluation of $H_\epsilon$.

%%%%%%%%%%%%%%%%%%%%%%%%%%%%%%%%%%%%%%%%%%%%%%%%%%%%%%%%%%%%%%%%%%%%%%%%%%%%%%%%
\subsection{Metrics}
\label{sec:metrics}

\begin{table}[t]
\centering
\caption{Metrics' definitions.}
        \begin{tabular}{@{}ll@{}}
            \toprule
            Metric & Definition \\
            \midrule
            Cost Ratio & $\langle \bar{W}, Y \rangle / \langle \bar{W}, \bar{Y} \rangle$ \\
            Generalized Cost Ratio & $\langle \bar{W}, Y^{\text{(rnd\_s)}} \rangle / \langle \bar{W}, \bar{Y}^{\text{(rnd\_s)}} \rangle$ \\
            Expanded Nodes & $\sum\nolimits_{n \in \mathcal{N}} E(n)$ \\
            Generalized Expanded Nodes & $\sum\nolimits_{n \in \mathcal{N}} E^{\text{(rnd\_s)}}(n)$ \\
            \bottomrule
        \end{tabular}
    \label{tab:metrics_def}
\end{table}

To measure the path prediction accuracy of the compared architectures, we use the \emph{cost ratio}, as in~\cite{black_box}. In order to account for cost-equivalent paths, we define the cost ratio as the ratio between the predicted path cost and the optimal path cost, according to the ground-truth costs $\bar{W}$. A cost ratio close to $1$ indicates that the system produces cost functions correctly generalizing on new maps. 

The cost ratio involves paths that start from the sample source and end in the sample target. Our goal, however, is to generate costs and heuristics that are source-agnostic. In principle, the cost function depends only on the image, while the heuristic also considers the target. The source point should not influence any of the two functions. To account for this behavior, we define the \emph{generalized cost ratio} as the cost ratio measured according to $Y^{\text{(rnd\_s)}}$, i.e., the path prediction from a random source point to the target. This new source is sampled uniformly from the valid sampling regions of the two datasets at each metric evaluation.

To measure the efficiency of the architectures, we simply count how many nodes have been expanded at the end of the A* execution. We refer to this metric as \emph{expanded nodes}. Also, we provide the \emph{generalized expanded nodes} metric to account for randomly sampled sources.
\Tab{\ref{tab:metrics_def}} collects the metrics' definitions.

%%%%%%%%%%%%%%%%%%%%%%%%%%%%%%%%%%%%%%%%%%%%%%%%%%%%%%%%%%%%%%%%%%%%%%%%%%%%%%%%
\subsection{Experiments}
\label{sec:experiments}

We compare Neural Weighted A* (\nwa) against the following baselines:

\myDescription{
    \myItem{\bba\ (\bbastar~\cite{black_box}).}{%
        A fully convolutional neural network computes $W$ from $I$. Then, a \bbastar\ module evaluates the shortest path $Y$. We follow the implementation of~\cite{black_box}, except for the Dijkstra algorithm, substituted by A* with admissible \cheb\ heuristic (\Eq{\ref{eq:cheb_h}}).
    }
    
    \myItem{\na\ (Neural A*~\cite{neural_a_star}).}{%
        A fully convolutional neural network computes $W$ from $I$, using $s$ and $t$ as additional input channels. Then, a Neural A* module~\cite{neural_a_star} evaluates the expanded nodes $E$. The non-admissible heuristic 
        \begin{equation}
            H_\text{NA*}(n) = D_C(n, t) + 0.001 \cdot D_E(n, t)
            \label{eq:non_adm_h}
        \end{equation}
        is used to speed up the search, as in~\cite{neural_a_star}. This heuristic is the weighted sum of the \cheb\ distance $D_C$ and the Euclidean distance $D_E$ between $n$ and $t$ in the grid. The non-admissibility arises from the fact that the scaling term $w_\text{min}$ is missing, differently from \Eq{\ref{eq:cheb_h}}. Despite reducing the expanded nodes, this heuristic adds a strong bias towards paths that move straight to the target, greatly penalizing the cost ratio.
    }
    
    \myItem{\admna\ (Admissible Neural A*).}{%
        We propose this architecture as a clone of the original \na\ architecture~\cite{neural_a_star}, but we substitute the non-admissible heuristic $H_\text{NA*}$ (\Eq{\ref{eq:non_adm_h}}) with the admissible \cheb\ heuristic $H_C$ (\Eq{\ref{eq:cheb_h}}). Our goal is to minimize the influence of the fixed, non-admissible heuristic $H_\text{NA*}$~\cite{neural_a_star} on the expanded nodes $E$. In this way, the numerical results reflect more accurately the neural predictions, ensuring a fair comparison with \nwa.
    }
    
    \myItem{\nsna\ (No-Source Neural A*).}{%
        This architecture is equal to \admna, except for the source channel, not included in the neural network input. Differently from \na\ and \admna, by hiding the information related to the source node location, we expect \nsna\ to exhibit no sensible performance downgrade between the cost ratio values and the generalized cost ratio values. The same holds for the expanded nodes and the generalized expanded nodes.
    }
}

%%%%%%%%%%%%%%%%%%%%%%%%%%%%%%%%%%%%%%%%%%%%%%%%%%%%%%%%%%%%%%%%%%%%%%%%%%%%%%%%
\subsection{Implementation details}
\label{sec:impl_details}

Each architecture uses convolutional layer blocks from ResNet18~\cite{resnet}, as in~\cite{black_box}, to transform tile-based images into cost or heuristic functions, encoded as single-channel tensors. We substitute the first convolution to adapt to the number of input channels, varying between $3$~(\bba), $4$~(\nsna, \nwa), and~$5$~(\na, \admna). In \bba, we perform average pooling to reduce the output channels to $1$. Then, to ensure that the weights are non-negative, we add a ReLU for Warcraft and a sigmoid for Pokémon. All the baselines involving Neural A* (\na, \admna, and \nsna), instead, include a $1 \times 1$ convolution followed by a sigmoid. Finally, in \nwa, the channel-reduction strategy depends on the neural network. The cost-predicting ResNet18 is followed by an average operation and a normalization between $w_{\text{min}} = 1$ and $w_{\text{max}} = 10$. The heuristic-predicting ResNet18 is followed by a $1 \times 1$ convolution and a normalization in the range $[0, 1]$. Each architecture trains with the Adam optimizer. The learning rate is equal to $0.001$. The batch size is $64$ for Warcraft and $16$ for Pokémon to account for GPU memory usage. The $\lambda$ parameter of the black-box solvers of \bba\ and \nwa\ is $20$. The $\tau$ parameter of the Neural A* solvers of \na, \admna, \nsna, and \nwa\ is set to the square root of the grid width, so $3.46$ for Warcraft and $4.47$ for Pokémon. In \Eq{\ref{eq:loss}}, we impose $\alpha = 1$ and $\beta = 0.1$ to bring the loss components to the same order of magnitude. We found that the training procedure is not affected by small deviations from the parameters described in this section. Finally, we detected sensible improvements in the efficiency when training \nwa\ with random $\epsilon$, as it expanded noticeably fewer nodes than training with fixed $\epsilon$. Since the other metrics do not exhibit sensible differences, we always refer to the test results of \nwa\ obtained after training with $\epsilon$ sampled from $[0, 9]$ at each $H_\epsilon$ evaluation.
%\Fig{\ref{fig:diff_eps_values}} shows the generalized expanded nodes metric (y-axis) as $\epsilon$ increases at testing time (x-axis) for the Warcraft dataset. We compare five \nwa\ architectures, each trained with a different $\epsilon$-related strategy. The first three are trained with $\epsilon = 1$, $4$, and $9$. The last two are trained with $\epsilon$ sampled uniformly from $[0, 9]$ and $[0, 15]$ at each $H_\epsilon$ evaluation. At testing time, we measure the generalized expanded nodes metric for several $\epsilon$ values. We see that the architectures trained with random $\epsilon$ expand noticeably fewer nodes than the fixed-$\epsilon$ counterparts. Since the same also holds for the Pokémon dataset, and the other metrics do not exhibit sensible differences, we always refer to the test results of \nwa\ obtained after training with $\epsilon$ sampled from $[0, 9]$.
To ensure the full reproducibility of our experiments, we share the source code.\footnote{\sourceurl}

\begin{comment}
\begin{figure}[t]
    \centering
    \includegraphics[width=.48\textwidth]{img/eps_train.pdf}
    \caption{Generalized expanded nodes on \nwa\ architectures trained with different $\epsilon$-related strategies on the Warcraft dataset.}
    \label{fig:diff_eps_values}
\end{figure}
\end{comment}

%%%%%%%%%%%%%%%%%%%%%%%%%%%%%%%%%%%%%%%%%%%%%%%%%%%%%%%%%%%%%%%%%%%%%%%%%%%%%%%%
\subsection{Results}
\label{sec:results}

\begin{table}[t]
    \centering
    \caption{Quantitative results on the Warcraft and Pokémon datasets.}
    \centering
    \begin{tabular}{@{}lc|cccc|cccc@{}}
\toprule
&& \multicolumn{4}{c}{Warcraft dataset} & \multicolumn{4}{|c}{Pokémon dataset} \\
Experiment & $\epsilon$ & \metrics & \metrics \\ \midrule
\bba & - & \meanstdbf{1.0}{0.0} & \meanstdbf{1.0}{0.0} & \meanstd{69.8}{0.16} & \meanstd{69.94}{0.42} & \meanstd{1.57}{0.03} & \meanstd{1.65}{0.03} & \meanstd{79.78}{1.02} & \meanstd{79.29}{0.6} \\
\na & - & \meanstd{1.29}{0.0} & \meanstd{1.41}{0.01} & \meanstdbf{9.81}{0.0} & \meanstdbf{9.82}{0.01} & \meanstd{2.15}{0.05} & \meanstd{2.57}{0.07} & \meanstdbf{15.02}{0.0} & \meanstdbf{14.64}{0.03} \\  
\admna & - & \meanstd{1.04}{0.0} & \meanstd{1.17}{0.01} & \meanstd{13.12}{0.06} & \meanstd{22.48}{0.37} & \meanstdbf{1.11}{0.02} & \meanstd{1.24}{0.02} & \meanstd{26.57}{1.89} & \meanstd{37.03}{1.63} \\  
\nsna & - & \meanstd{1.12}{0.01} & \meanstd{1.11}{0.0} & \meanstd{21.19}{0.47} & \meanstd{21.19}{0.45} & \meanstd{1.16}{0.02} & \meanstdbf{1.17}{0.02} & \meanstd{39.76}{1.71} & \meanstd{39.44}{1.53} \\ 
\midrule 
\nwa & $0.0$ & \meanstdbf{1.0}{0.0} & \meanstdbf{1.0}{0.0} & \meanstd{68.42}{1.2} & \meanstd{69.04}{1.18} & \meanstd{1.06}{0.03} & \meanstd{1.05}{0.02} & \meanstd{124.06}{1.89} & \meanstd{121.2}{1.9} \\ 
\nwa & $1.0$ & \meanstd{1.01}{0.0} & \meanstd{1.01}{0.0} & \meanstd{49.54}{1.21} & \meanstd{50.57}{1.3} & \meanstdbf{1.03}{0.01} & \meanstdbf{1.03}{0.01} & \meanstd{80.21}{1.73} & \meanstd{78.43}{1.46} \\  
\nwa & $4.0$ & \meanstd{1.03}{0.0} & \meanstd{1.03}{0.0} & \meanstd{26.61}{0.29} & \meanstd{27.15}{0.17} & \meanstd{1.08}{0.04} & \meanstd{1.06}{0.02} & \meanstd{56.4}{1.92} & \meanstd{56.21}{1.5}\\  
\nwa & $9.0$ & \meanstd{1.1}{0.01} & \meanstd{1.09}{0.01} & \meanstd{14.2}{0.8} & \meanstd{14.47}{0.8} & \meanstd{1.22}{0.12} & \meanstd{1.14}{0.04} & \meanstd{31.54}{2.35} & \meanstd{31.04}{2.73} \\  
\nwa & $11.0$ & \meanstd{1.13}{0.03} & \meanstd{1.11}{0.01} & \meanstd{12.74}{0.88} & \meanstd{13.0}{0.89} & \meanstd{1.22}{0.11} & \meanstd{1.24}{0.08} & \meanstd{23.37}{0.85} & \meanstd{23.06}{0.98} \\  
\nwa & $14.0$ & \meanstd{1.15}{0.03} & \meanstd{1.13}{0.02} & \meanstdbf{11.94}{0.5} & \meanstdbf{12.12}{0.49} & \meanstd{1.22}{0.09} & \meanstd{1.35}{0.13} & \meanstdbf{20.09}{0.78} & \meanstdbf{19.84}{0.92} \\  
\bottomrule
\end{tabular}
    \label{tab:quantitative_results}
\end{table}

We collect the quantitative results of our experiments in \Tab{\ref{tab:quantitative_results}}. The table is split into four sections, two for each dataset. Each section contains either baseline experiments or \nwa-related experiments measured on the same \nwa\ architecture fixing different $\epsilon$ values at testing time. We comment on the experiments'~results by answering the following three questions:

\begin{figure*}[t]
    \centering
    \includegraphics[width=\textwidth]{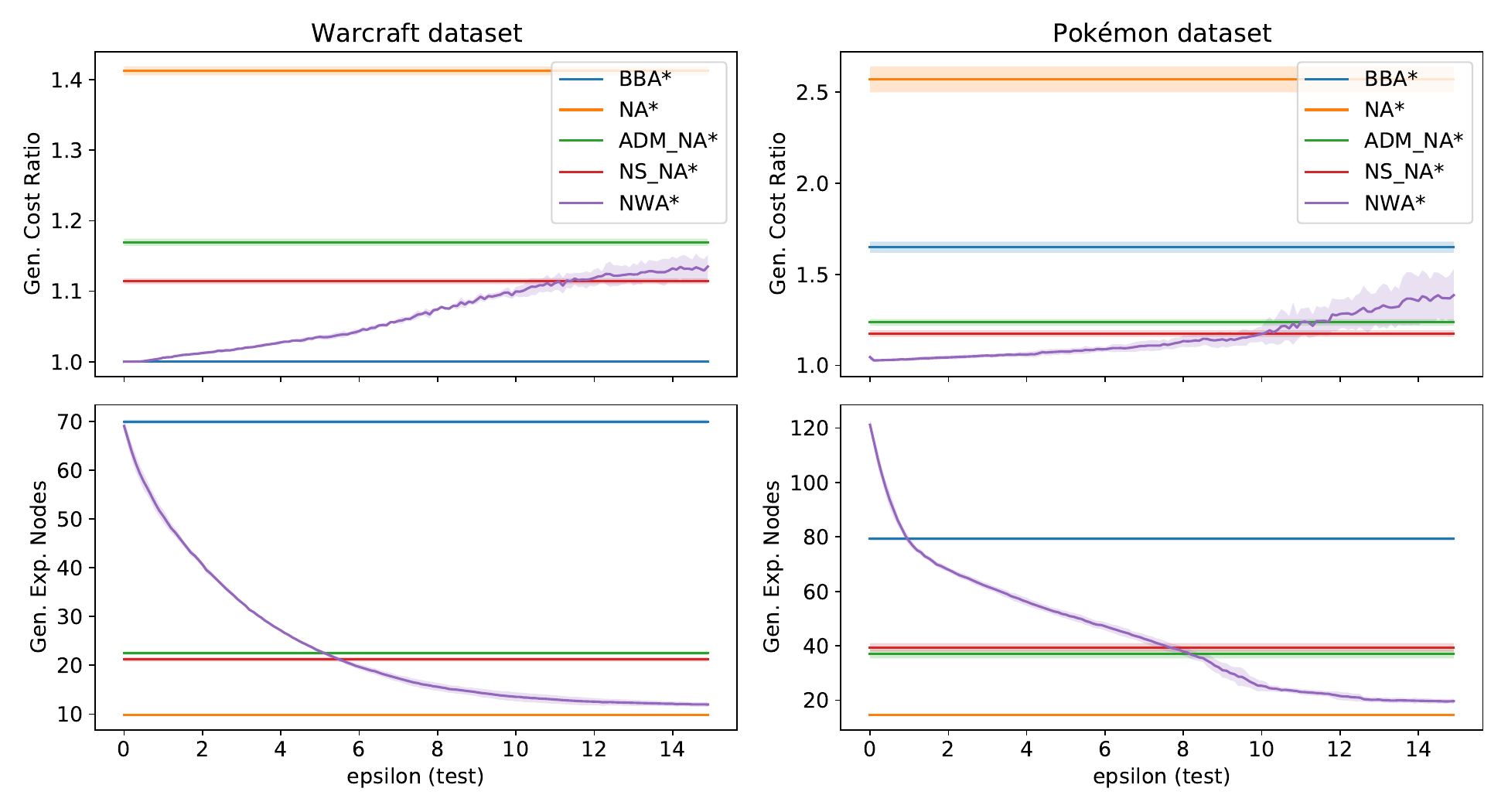}
    \caption{Comparison between generalized cost ratio and generalized expanded nodes across the experiments for several $\epsilon$ values (mean $\pm$ std over five restarts). 
    %For both metrics, the lower the better. 
    For low $\epsilon$, \nwa\ is the most accurate model, while for higher $\epsilon$, it is the most efficient. The only exception is \na, which is barely faster but much less reliable in terms of cost ratio.}
    \label{fig:metrics_comparison}
\end{figure*}

\myDescription{
    \myItem{Does NWA* learn to predict cost functions correctly?}{%
        %The (generalized) cost ratio metric measures the path prediction accuracy of the architectures. 
        By observing Table~\ref{tab:quantitative_results}, \nwa\ reaches a nearly perfect cost ratio on both datasets for a considerable range of $\epsilon$ values. This was expected for $\epsilon \approx 0$, but the (generalized) cost ratio remains very low for $\epsilon$ up to $4$, which is an excellent result considering the corresponding expanded nodes speedup. In Warcraft, \nwa\ behaves as \bba\ cost-ratio-wise, while, in Pokémon, it outperforms all the baselines. Since, by setting $\epsilon$ to low values, the path predictions do not take into account $H_\epsilon$, the positive cost-ratio-related performance implies that \nwa\ learned to predict cost functions that make A* return paths close to the ground-truth.
    }
    \myItem{Does NWA* learn to predict heuristic functions correctly?}{%
        %The (generalized) expanded nodes metric measures the efficiency of A* run on $W$ and $H_\epsilon$. 
        By setting $\epsilon \gg 0$, $H_\epsilon$ drives the search. A* converges faster, but it may return suboptimal paths. Therefore, we expect a small penalty on cost ratios, but a noticeable decrease in the node expansions. Again, the empirical results confirm this trend on both datasets. For $\epsilon = 14$, we outperform all the baselines in terms of generalized node expansions. The only exception is \na, which expands fewer nodes, but exhibits an extremely higher cost ratio. %This is due to the non-admissible heuristic $H_\text{NA*}$ (\Eq{\ref{eq:non_adm_h}}) that drives the search, as in~\cite{neural_a_star}, adding a strong bias towards straight-line paths. 
        \nwa\ trades off few node expansions to be much more reliable than \na\ in terms of path predictions.
    }
    \myItem{Can NWA* trade off planning accuracy for efficiency?}{%
        By setting $\epsilon$ close to $0$, \nwa\ behaves accurately (low cost ratio, high expanded nodes). By increasing $\epsilon$, \nwa\ becomes more efficient (higher cost ratio, lower expanded nodes). 
        %This trend is confirmed by the numerical results of \Tab{\ref{tab:quantitative_results}}. However, to
        To visually illustrate the extent of the tradeoff capabilities of \nwa, we plot in \Fig{\ref{fig:metrics_comparison}} the generalized metrics (y-axis) for all the experiments with respect to several $\epsilon$ values (x-axis). Since the baseline architectures do not depend on $\epsilon$, their behavior is plotted as a horizontal line for comparison. \nwa, on the other hand, smoothly interpolates between the accuracy of \bba\ and the efficiency of \na-related architectures, offering to the user the possibility of finely tuning $\epsilon$ to the desired planning behavior, from the most accurate to the most efficient. 
    }
}
\section{Conclusions}

With Neural Weighted A*, we propose a differentiable, anytime shortest path solver able to learn graph costs and heuristics for planning on raw image inputs. The system trains with direct supervision on planning examples, making data labeling cheap. Unlike any similar data-driven planner, we can choose to return the optimal solution or to trade off accuracy for convergence speed by tuning a single, real-valued parameter, even at runtime. We guarantee the solution suboptimality to be constrained within a linear bound proportional to the tradeoff parameter. We experimentally test the validity of our claims on two tile-based datasets. By inspecting the numerical results, we see that Neural Weighted A* consistently outperforms the accuracy and the efficiency of the previous works, obtaining, in a single architecture, the best of the two worlds.

%%%%%%%%%%%%%%%%%%%%%%%%%%%%%%%%%%%%%%%%%%%%%%%%%%%%%%%%%%%%%%%%%%%%%%%%%%%%%%%%
%\addtolength{\textheight}{-12cm}
\bibliographystyle{splncs04} 
\bibliography{references}
\end{document}